\pdfoutput=1

\documentclass[11pt]{article}

\usepackage[final]{acl}

\usepackage{times}
\usepackage{latexsym}
\usepackage{amsmath}
\usepackage{booktabs}
\usepackage{multirow} 
\usepackage{bm} 
\usepackage[T1]{fontenc}
\usepackage{amssymb}  
\usepackage[table,dvipsnames]{xcolor}

\usepackage[utf8]{inputenc}
\usepackage{diagbox}

\usepackage{microtype}

\usepackage{inconsolata}

\usepackage{graphicx}
\usepackage{ulem}
\usepackage{xcolor}
\definecolor{cctgreen}{RGB}{0,176,80}
\title{LLM-Driven Completeness and Consistency Evaluation for Cultural Heritage Data Augmentation in Cross-Modal Retrieval}

\author{
  \textbf{Jian Zhang\textsuperscript{1$\dagger$}}
  \textbf{Junyi Guo\textsuperscript{1$\dagger$}},
  \textbf{Junyi Yuan\textsuperscript{1}},
  \textbf{Huanda Lu\textsuperscript{2}},
\\
  \textbf{Yanlin Zhou\textsuperscript{3}},
  \textbf{Fangyu Wu\textsuperscript{1$^\ast$}},
  \textbf{Qiufeng Wang\textsuperscript{1}},
  \textbf{Dongming Lu\textsuperscript{4}}
\\
\\
  \textsuperscript{1}Xi'an Jiaotong-Liverpool University,
  \textsuperscript{2}NingboTech University,\\
  \textsuperscript{3}Dunhuang Academy,
  \textsuperscript{4}Zhejiang University
\\
  \small{
    \textbf{Correspondence:} \href{mailto:fangyu.wu02@xjtlu.edu.cn}{fangyu.wu02@xjtlu.edu.cn}
  }
}

\begin{document}
\maketitle
\renewcommand{\thefootnote}{\fnsymbol{footnote}}
\footnotetext[2]{Equal contribution.}
\footnotetext[1]{Corresponding author.}
\begin{abstract}

Cross-modal retrieval is essential for interpreting cultural heritage data, but its effectiveness is often limited by incomplete or inconsistent textual descriptions, caused by historical data loss and the high cost of expert annotation. While large language models (LLMs) offer a promising solution by enriching textual descriptions, their outputs frequently suffer from hallucinations or miss visually grounded details. To address these challenges, we propose $C^3$, a data augmentation framework that enhances cross-modal retrieval performance by improving the completeness and consistency of LLM-generated descriptions. $C^3$ introduces a completeness evaluation module to assess semantic coverage using both visual cues and language-model outputs. Furthermore, to mitigate factual inconsistencies, we formulate a Markov Decision Process to supervise Chain-of-Thought reasoning, guiding consistency evaluation through adaptive query control. Experiments on the cultural heritage datasets CulTi and TimeTravel, as well as on general benchmarks MSCOCO and Flickr30K, demonstrate that $C^3$ achieves state-of-the-art performance in both fine-tuned and zero-shot settings. The code of this paper is available at \url{https://github.com/JianZhang24/C-3}.

\end{abstract}

\section{Introduction}
Cultural heritage reflects the historical, artistic, and social dimensions of human civilization across regions and periods ~\cite{nilson2018cultural}.
As traditional cultural heritage often comprises abstract images or patterns, textual descriptions are crucial for connecting visual data to meaningful cultural interpretations.
In this paper, we focus on cross-modal retrieval tasks that enable effective matching and interpretation of cultural heritage data. This capability is fundamental for real-world applications such as digital preservation and interactive museum systems.

Cross-modal retrieval, which aims to retrieve relevant samples in one modality given a query from another, has been extensively studied on general datasets such as MSCOCO~\cite{mscoco} and Flickr30K~\cite{30k}. Recent advances have achieved significant improvements by enhancing representation learning and designing fine-grained matching strategies ~\cite{radford2021learning, huang2024cross, yang2024continual}.
\begin{figure}[t]
  \centering
  \includegraphics[width=0.95\linewidth]{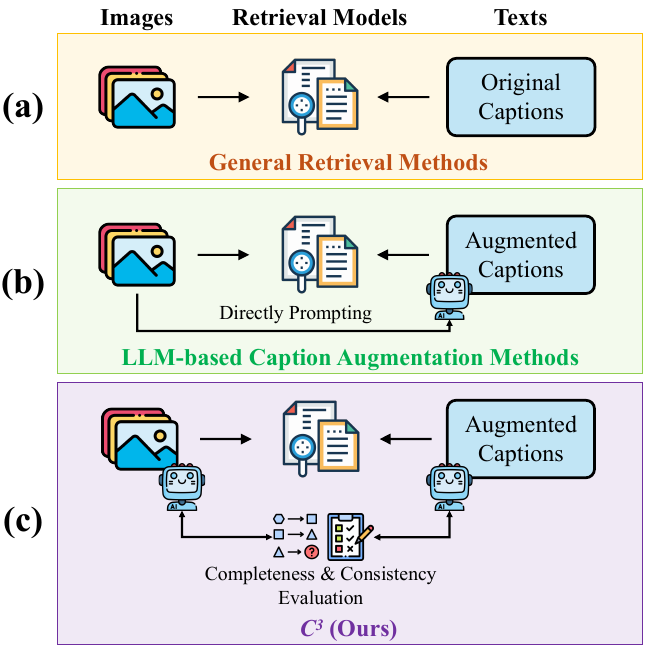}
  \caption{Illustration of our $C^3$ with the general retrieval methods and the LLM-based caption augmentation methods.}
  \label{fig:example}
\end{figure}
Nevertheless, cross-modal retrieval performance can degrade significantly when textual descriptions lack accuracy, completeness, or sufficient detail~\cite{ma2025multi,wang2025cross,sogi2024object}. This issue is even more pronounced in the cultural heritage domain, where historical data loss leads to incomplete records, 
and producing reliable annotations often requires interdisciplinary expertise. The resulting sparsity and inconsistency in textual descriptions substantially hinder the effectiveness of multimodal retrieval in this context.


In recent years, the rapid development of large language models (LLMs)~\cite{guo2025deepseek,hurst2024gpt,qwen2.5} has led to significant advancements in enhancing textual descriptions~\cite{wan2024tnt,wu2024improving,hu2024llm}. Inspired by these advances, we explore LLM-based text augmentation methods to address the limited completeness and fidelity of textual description, thereby improving cross-modal retrieval performance. To achieve this goal, two key challenges arise: (1) how to ensure that the augmented textual annotations comprehensively capture the relevant image content; (2) how to ensure the factual accuracy of generated descriptions by addressing hallucinations introduced by LLMs.


In this paper, we propose $C^3$, a LLM-driven data augmentation framework designed to enhance \textbf{C}ross-modal retrieval in cultural heritage domains by improving the \textbf{C}ompleteness and \textbf{C}onsistency of textual descriptions. As shown in Fig. \ref{fig:example}(a), general retrieval methods rely on feature alignment between the image and the original caption. LLM-augmented approaches (Fig.~\ref{fig:example}(b)) instead enhance the original caption, but often overlook whether each generated detail is grounded in visual evidence, leading to hallucinated or incomplete descriptions. To address this limitation, $C^3$ enhances supervision by augmenting 
descriptions and explicitly validating their completeness and factual correctness to reduce hallucination, thereby improving retrieval performance.

For completeness, $C^3$ introduces a bidirectional coverage attention evaluation approach that ensures both visual and textual information are mutually verified, establishing a cross-check mechanism to capture all relevant attributes. We further implement a coverage-based scoring method to measure how well the generated captions reflect key semantic content. This design enables our model to check that all relevant visual and textual information is captured and aligned. To improve consistency, we propose a Chain-of-Thought prompting strategy and supervise it using a Markov decision process. This supervision guides each CoT reasoning step and mitigates hallucinations by ensuring that generated textual description is grounded in visual evidence. Together, these innovations jointly enable $C^3$ to generate enhanced captions that are both semantically comprehensive and factually reliable, leading to improved cross-modal retrieval performance. 

In summary, our contribution has four folds:
\begin{itemize}

\item We propose $C^3$, a large language model-driven data augmentation framework for cross-modal retrieval that validates generated descriptions for completeness and factual consistency. Rather than modifying retrieval model architectures, $C^3$ improves performance by enhancing the quality and reliability of textual supervision, enabling robust retrieval under incomplete text description.

\item We introduce a novel completeness evaluation module via bidirectional coverage attention evaluation that integrates attention mechanisms with large language models, enabling comprehensive cross-verification between visual and textual modalities.
\item We design a Chain-of-Thought prompting strategy supervised by a Markov decision process, effectively reducing hallucinations and improving the consistency of augmented textual description.
\item Extensive experiments show that our method achieves state-of-the-art results on cultural heritage datasets CulTi and TimeTravel, as well as on general benchmarks MSCOCO and Flickr30K under both fine-tuning and zero-shot settings.
\end{itemize}

\section{Related Work}
\subsection{Cross-Modal Retrieval}
Existing cross-modal retrieval methods are typically divided into global and local alignment approaches.
Global alignment methods such as CLIP~\cite{radford2021learning} and ALIGN~\cite{jia2021scaling} employ contrastive learning to align image-text pairs at the representation level. Chinese-CLIP~\cite{yang2022chinese} extends these approaches specifically to Chinese datasets, addressing linguistic and cultural specificities. In comparison, local alignment methods focus on fine-grained associations between image regions and corresponding textual elements. 
For example, LexVLA~\cite{li2024unified} improves interpretability and reduces false matches by introducing an overuse penalty mechanism.
Similarly, GOAL~\cite{choi2025goal} leverages the segmentation model to achieve more accurate local alignment. 
Recent studies~\cite{chen2023rethinking,liu2024flickr30k,pan2023fine,wu2024discriminative} have shown that incomplete or under-specified textual descriptions pose challenges for cross-modal retrieval. This issue is further exacerbated in cultural heritage domain, which typically lack detailed and accurate descriptions. In this paper, we explore the caption augmentation for improving cross-modal retrieval performance.

\subsection{Large Language Models-based Data Augmentation}
Recent advances in LLMs have enhanced data augmentation by enabling the generation of high-quality textual descriptions, which helps address annotation scarcity and improves generalization ~\cite{wulotlip,fan2023improving}. PhiloGPT~\cite{zhang-etal-2024-philogpt} and DAR~\cite{song2024enhancing} demonstrate the use of LLM-based data augmentation in specific domains to address annotation scarcity. 
However, LLMs often exhibit “hallucination”, producing content that appears plausible but lacks a factual basis~\cite{10.1145/3703155}, leading to inaccurate or incomplete descriptions. To address this issue,
\citet{alizadeh2023open} propose a reference-based comparison with human annotations to quantify output quality and filter unreliable samples. DoAug~\cite{wang2025diversity} addresses hallucination by fine-tuning models on high-quality, validated datasets. While such methods have demonstrated effectiveness in reducing hallucination, they often incur high labor costs and risk overfitting. 
In this paper, we aim to improve both the completeness and consistency of LLM-augmented textual descriptions, while avoiding costly human supervision and retaining scalability across domains.

\begin{figure*}
    \centering
    \includegraphics[width=0.9\linewidth]{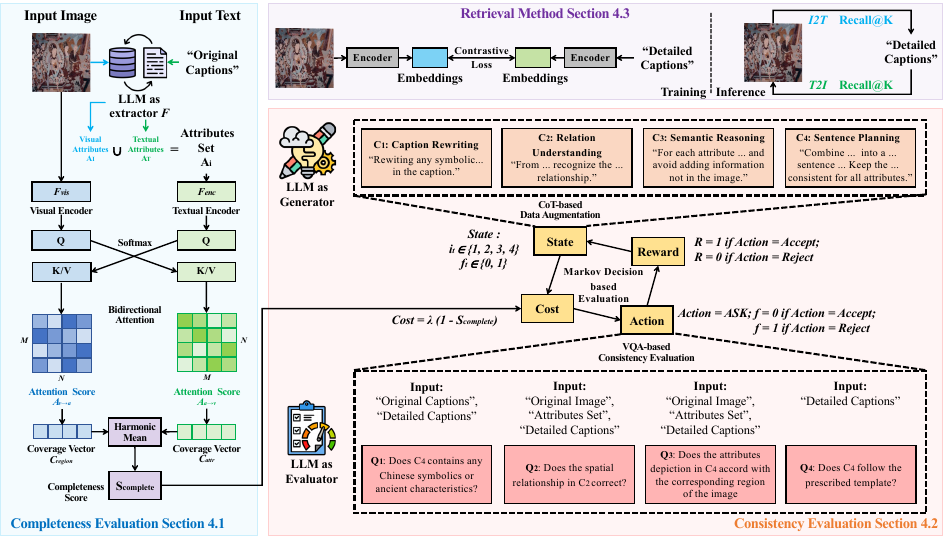}
    \caption{Overview of the proposed $C^3$ framework. The pipeline first verifies image-text attributes extracted by an LLM with bidirectional attention and coverage scoring, then augments captions through a CoT framework and consistency evaluation process. Detailed captions are used to fine-tune a CLIP-based retrieval model for improved image-text aligning.}
    \label{fig2}
\end{figure*}

\section{Motivation}
Recent advances in LLM-based data augmentation have motivated us to explore their potential for improving cross-modal retrieval in the cultural heritage domain. In this section, we first investigate the necessity of evaluating the completeness of augmented captions through LLM in Section~3.1, and then discuss the role of Markov decision in enhancing the consistency of CoT-based caption augmentation in Section 3.2.

\subsection{Completeness Evaluation for LLM-Augmented Descriptions}

In cultural heritage retrieval, LLMs offer a promising solution to incomplete or partially informative descriptions by enriching them with additional semantic content for improved cross-modal alignment. 
However, the effectiveness of augmentation relies on generating descriptions that comprehensively capture all relevant visual attributes. One intuitive strategy is to use VLLMs to extract such attributes and assess whether they are sufficiently reflected in the text \cite{wang2023visionllm, wu2024visionllm}. Although VLLMs are capable of generating semantically rich descriptions, their reliance on language priors often leads to incomplete coverage of visually grounded attributes. Conversely, attention-based methods offer more grounded supervision but often fail to capture rare or fine-grained visual elements, leading to incomplete coverage.


To address these limitations, we draw inspiration from coverage-based attention mechanisms developed for text summarization~\cite{see2017get,tu2016modeling}, which explicitly track how thoroughly source content is represented during generation. Building on this idea, we propose a completeness evaluation framework that combines VLLM-driven attribute extraction with bidirectional coverage attention evaluation. As discussed in Section 4.2, this design enables more accurate assessment of LLM-augmented descriptions and leads to better alignment between visual content and textual representations in cross-modal retrieval.


\subsection{Consistency Enhancement for CoT-Based LLM Descriptions}

Chain-of-Thought (CoT) prompting provides a structured way for LLMs to generate semantically rich and coherent captions, especially in contexts requiring contextual understanding. However, this step-by-step reasoning format can also expose a key limitation of LLMs: the frequent introduction of hallucinated or inaccurate content. Such hallucinations typically occur when generating contextually rich and culturally specific captions, leading to incorrect or nonexistent symbolic attributes being described. For example, a caption may incorrectly describe symbolic details not actually present in the image, severely impacting retrieval accuracy. Addressing these factual consistency issues is therefore critical to improving the accuracy of CoT-based caption augmentation.

To mitigate this issue, we aim to introduce a mechanism that can maintain logical consistency and factual alignment throughout the reasoning process. Existing approaches \cite{shum2023automatic, li2024cot} often treat each generation step independently, which leads to semantic drift and factual errors over time. To address this, we explore the use of Markov Decision Processes (MDPs), which are effective for modeling sequential dependencies. By aligning CoT's step-by-step reasoning with the structure of MDPs, we guide the generation process more coherently and reduce hallucinations, which in turn improves factual consistency and enhances retrieval performance.


\section{Methods}
We propose the $C^3$ framework, which leverages large language models to generate complete and consistent textual descriptions, thereby enhancing cross-modal retrieval performance. The overall workflow is shown in Fig. \ref{fig2}. 

\subsection{Completeness Evaluation}
Given an image–text pair $(T_i, I_i)$, the proposed $C^3$ first extracts attribute-level features from both modalities. It then evaluates their cross-modal correspondence and measures coverage completeness to ensure semantic alignment.
\subsubsection{Attribute Extraction}
For each modality $x \in {I_i, T_i}$, we utilize a pretrained multimodal large language model $F$ to extract a set of semantic attributes:
\begin{equation}
A_x = F(x) = {a^x_1, a^x_2, \dots, a^x_{n_x}}
,\end{equation}
where $a^x_j$ denotes the $j$-th attribute extracted from modality $x$, and $n_x$ is the total number of attributes identified in $x$.
We then merge both attribute sets into an integrated attribute pool:
\begin{equation}
    A_i = A_{x=I_i} \cup A_{x=T_i}.
\end{equation}
${A}_i$ provides a concise basis for subsequent completeness evaluation.

\subsubsection{Bidirectional Coverage Attention Evaluation} 
To quantitatively evaluate the completeness of attribute coverage relative to image regions, we propose a bidirectional coverage attention evaluation. Unlike existing approaches that consider coverage from only one modality, our method simultaneously evaluates both visual-to-textual and textual-to-visual coverage.
First, visual embeddings $V_{I_i}$ for image $I_i$ are computed using a pretrained vision encoder $F_{\text{vis}}$ to obtain semantic embeddings:
\begin{equation}
    V_{I_i} = F_{\mathrm{vis}}(I_i) \in \mathbb{R}^{N \times d_v},
\end{equation}
where $N$ is the number of spatial regions in the visual feature map and $d_v$ is the visual embedding dimension.

Next, attribute embeddings $E({A}_i)$ for ${A}_i$ are generated via a pretrained textual encoder $F_{\text{enc}}$:
\begin{equation}
    E({A}_i) = F_{\mathrm{enc}}({A}_i) \in \mathbb{R}^{M \times d_t},
\end{equation}
where $M$ represents the number of attributes and $d_t$ denotes the dimension of textual embeddings.
To jointly evaluate how comprehensively the visual attributes cover each spatial region and how thoroughly each attribute is grounded in visual evidence, we apply the visual regions in the image as queries while attribute embeddings as keys/values:
\begin{equation}
\label{eq:vis2attr_att} 
A_{\text{v}\rightarrow\text{a}}
     = \mathrm{softmax}\!\Bigl(
       \tfrac{V_{I_i}W^{Q}\big(E({A}_i)W^{K}\big)^\top}{\sqrt{d_k}}
       \Bigr)
       \in \mathbb{R}^{N\times M}
,\end{equation}
where $W^{Q}\!\in\!\mathbb{R}^{d_v\times d_k}$ and $W^{K}\!\in\!\mathbb{R}^{d_t\times d_k}$ are learnable projections.

Then, we apply attributes as queries while the visual tokens as keys/values:
\begin{equation}
\label{eq:attr2vis_att} 
A_{\text{a}\rightarrow\text{v}}
     = \mathrm{softmax}\!\Bigl(
       \tfrac{E({A}_i)W^{Q'}\big(V_{I_i}W^{K'}\big)^\top}{\sqrt{d_k}}
       \Bigr)
       \in \mathbb{R}^{M\times N}
,\end{equation}
where $W^{Q'},W^{K'}\!\in\!\mathbb{R}^{d_t\times d_k}$ are another sets of learnable projections.

We calculate the maximum attention along the attended dimension yields two coverage vectors:
\begin{equation}
\label{eq:cov_region} 
C^{\text{region}}_j = \max_{1\le m\le M}\!\bigl[A_{\text{v}\rightarrow\text{a}}\bigr]_{j,m},
\quad j=1,\dots,N,
\end{equation}
\begin{equation}
\label{eq:cov_attr} 
C^{\text{attr}}_m   = \max_{1\le j\le N}\!\bigl[A_{\text{a}\rightarrow\text{v}}\bigr]_{m,j},
\quad m=1,\dots,M.
\end{equation}

\paragraph{Unified Completeness Score.} 
We compute the mean coverage of image regions, $\bar{c}_{\text{region}}$, and the mean coverage of attributes, $\bar{c}_{\text{attr}}$, defined respectively as:
\begin{equation}
\bar{c}_{\text{region}} = \frac{1}{N}\sum_{j=1}^{N} C^{\text{region}}_j,\qquad
\bar{c}_{\text{attr}}   = \frac{1}{M}\sum_{m=1}^{M} C^{\text{attr}}_m
.\end{equation}
These two averages are then combined using the harmonic mean to ensure that high completeness scores require balanced coverage in both directions. The harmonic mean specifically penalizes cases where either visual-region or attribute coverage is weak, thereby enforcing a robust evaluation.
\begin{equation}
\label{eq:S_complete} 
S_{\text{complete}}
  = \frac{2\,\bar{c}_{\text{region}}\,\bar{c}_{\text{attr}}}
         {\bar{c}_{\text{region}}+\bar{c}_{\text{attr}}}
  \;\in [0,1].
\end{equation}
The unified completeness score $S_{\text{complete}}$ not only quantifies alignment quality but also guides decisions in subsequent consistency evaluation, dynamically modulating evaluation thoroughness based on coverage reliability.

\subsection{Consistency Evaluation}
\subsubsection{CoT-based Data Augmentation}
To enhance the consistency of cultural-heritage captions while reducing hallucinations, we propose a structured chain-of-thought (CoT) based data augmentation framework. Specifically, we employ a CoT strategy with four stages, denoted as
\(\{\mathrm{C}_1,\mathrm{C}_2,\mathrm{C}_3,\mathrm{C}_4\}\), to expand cultural‑heritage captions while guarding against hallucination. In $C^3$, we use carefully designed prompts corresponding to each phase as shown in Fig. \ref{fig2}. By progressing from \(\mathrm{C}_1\) to \(\mathrm{C}_4\), our approach incrementally enhances the clarity, factuality, and consistency of the generated captions in a step-wise manner. 

\paragraph{Caption rewriting \(\mathrm{C}_1\)} Convert historical or symbolic expressions in the original textual descriptions into clear language to improve readability.

\paragraph{Relation Understanding \(\mathrm{C}_2\)} 
Identify spatial and action-level relationships between attributes based on visual evidence, ensuring that semantic connections are grounded in the image.

\paragraph{Cultural semantic reasoning \(\mathrm{C}_3\)} Construct a reasoning chain for each attribute and convert it into a factual clause, constraining the output format and length without introducing external knowledge.

\paragraph{Sentence planning \(\mathrm{C}_4\)} Apply a template to ensure the accuracy of the augmented descriptions.

\subsubsection{Markov Decision-based Consistency Evaluation}
To ensure consistency of the captions generated at each CoT augmentation stage, we frame the entire evaluation sequence as a Markov Decision Process (MDP) that supervises which questions are asked, when, and how strictly they are evaluated. This formulation offers two key advantages. First, traditional VQA‑only filters apply a fixed list of queries regardless of sample difficulty. This design often leads to hallucinations when low‑confidence answers are chained together. By embedding query issuance in an MDP with query cost, our framework minimizes the hallucination risk. Second, the CoT stages follow a set order: later steps build on the earlier ones. The MDP’s state explicitly tracks the current stage, ensuring that large‑language models ask questions that respect this temporal dependency instead of prematurely probing later facts. 

At each evaluation step $t$, the VLLM receives the caption generated by the CoT process. The corresponding state in the MDP process is defined as $s_t = (i_t, f_t)$, where $i_t \in \left\{1,2,3,4\right\}$ indicates the current CoT stage under evaluation, and $f_t \in \left\{0,1\right\}$ is set to 1 if any question receives a negative (“No”) response, signaling regeneration. The MDP is initialized with $i_1 = 1$ and $f_1 = 0$.

The module selects one of three actions at each state. For every stage C$_t$, we design a series of binary verification questions for each generation stage and integrate these queries as \textbf{ASK} actions within the MDP to govern the evaluation flow. The \textbf{ASK} action poses the next verification question. If the answer is positive, the VLLM proceeds to the next stage with $f_{t} = 0$ remaining unchanged. If the answer is negative, it sets $f_{t+1} = 1$, triggering regeneration. The \textbf{ACCEPT} action finalizes and retains the caption; it is allowed only when all stages ($C_1 \rightarrow C_4$) have passed verification with $i_t = 4$ and $f_t = 0$. The \textbf{REJECT} action prompts the model to regenerate the caption and restart the questioning process when a negative answer is received, indicated by $f_t = 1$.

The reward function balances evaluation thoroughness against query expense. Specifically, the ASK cost is defined as $Cost(s_t) = 1 - S_{\mathrm{complete}}$, where $S_{\mathrm{complete}} \in [0,1]$ measures current completeness. A binary reward is assigned only when all evaluation steps are successfully passed (\textbf{ACCEPT} after all stages); otherwise, any failed step triggers regeneration (\textbf{REJECT}) and results in zero reward. This MDP formulation explicitly manages the evaluation sequence, encouraging precise yet efficient question allocation. By dynamically adjusting evaluation thoroughness according to caption alignment quality, it effectively reduces hallucinations and ensures accurate caption generation. We denote the augmented text as $T_i^{aug}$, while the final retrieval target is denoted as $(T_i^{aug}, I_i)$.




\subsection{Contrastive Learning for Cross-modal Retrieval}
Based on the augmented textual descriptions from previous stages, we fine-tune a Chinese-CLIP model to enhance bidirectional image-text retrieval performance. 
Specifically, for each pair $(T_i^{aug}, I_i)$, the textual element \(T_i^{aug}\) is passed through pre-trained text encoder to obtain its feature embedding \(v_i^{T}\), whereas the visual element \(I_i\) is processed via a pretrained vision encoder to derive its feature embedding \(v^{I}_i\).
The resulting feature vectors \(v_i^{T}\) and \(v^{I}_i\) are projected into a shared multimodal embedding space. \(S(v_i^{T}, v^{I}_i)\) denotes the cosine similarity score between the text and image embeddings.
To better capture the bidirectional nature of image–text associations, a dual-objective formulation is adopted. For the image-to-text retrieval task, the training objective is formulated as:
\begin{equation}
L_{i2t}
= -\frac{1}{Z}
\sum_{i=1}^{Z}
\log
\frac{\exp\bigl(S(v^{I}_{i},\,v_i^{T})/\gamma\bigr)}
{\displaystyle\sum_{k=1}^{Z}\exp\bigl(S(v^{I}_{i},\,v_k^{T})/\gamma\bigr)},
\end{equation}
where $Z$ is the total number of image-text pairs, $\gamma$ is a learnable temperature parameter, and $i\neq k$. Similarly, the loss for the text-to-image retrieval task, $L_{t2i}$, is defined in a structurally symmetric form by interchanging the image and text representations. The overall training loss is formulated as $L = (L_{i2t} + L_{t2i})/2$. By minimizing the combined contrastive loss \(L\), the model aligns image and text embeddings in the joint multi-modal space.







\begin{table*}[ht]
\centering
\small
\renewcommand{\arraystretch}{0.9}
\setlength{\tabcolsep}{3pt}
\caption{Performance of $C^3$ on MSCOCO and Flickr30K datasets in zero-shot setting. Best results are in \textbf{bold}.}
\begin{tabular}{c|ccc|ccc|ccc|ccc}
\toprule
\multirow{2}{*}{} & \multicolumn{6}{c|}{\textbf{MSCOCO}} & \multicolumn{6}{c}{\textbf{Flickr30K}} \\
 {\centering \textbf{Methods}}& \multicolumn{3}{c|}{Image-to-Text} & \multicolumn{3}{c|}{Text-to-Image} & \multicolumn{3}{c|}{Image-to-Text} & \multicolumn{3}{c}{Text-to-Image} \\ 
 & R@1 & R@5 & R@10 & R@1 & R@5 & R@10 & R@1 & R@5 & R@10 & R@1 & R@5 & R@10 \\
\midrule

SCAN~\cite{lee2018stacked} & 50.4 & 82.2 & 90.0 & 38.6 & 69.3 & 80.4 & 67.4 & 90.3 & 95.8 & 48.6 & 77.7 & 85.2 \\
ViSTA~\cite{cheng2022vista} & 68.9 & 90.1 & 95.4 & 52.6 & 79.6 & 87.6 & 89.5 & 98.4 & 99.6 & 75.8 & 94.2 & 96.9 \\
COTS~\cite{lu2022cots} & 69.0 & 90.4 & 94.9 & 52.4 & 79.0 & 86.9 & 90.6 & 98.7 & 99.7 & 76.5 & 93.9 & 96.6 \\
LightningDoT~\cite{sun2021lightningdot} & 64.6 & 87.6 & 93.5 & 50.3 & 78.7 & 87.5 & 86.5 & 97.5 & 98.9 & 72.6 & 93.1 & 96.1 \\
LexLIP~\cite{luo2023lexlip} & 70.2 & 90.7 & 95.2 & 53.2 & 79.1 & 86.7 & \textbf{91.4} & \textbf{99.2} & \textbf{99.7} & \textbf{78.4} & \textbf{94.6} & \textbf{97.1} \\
CLIP~\cite{radford2021learning} & 51.8 & 76.8 & 84.3 & 32.7 & 57.7 & 68.2 & 44.1 & 68.2 & 77.0 & 24.7 & 45.1 & 54.6 \\
Long-CLIP~\cite{zhang2024long} & 62.8 & 85.1 & 91.2 & 46.3 & 70.8 & 79.8 & 53.4 & 77.5 & 85.3 & 41.2 & 64.1 & 72.6 \\
\midrule
\rowcolor{Magenta!5}\textbf{\bm{$C^3$}} (Ours) & \textbf{86.7} & \textbf{97.5} & \textbf{98.9} & \textbf{85.8} & \textbf{97.4} & \textbf{99.1} & 84.5 & 97.0 & 99.0 & 72.9 & 92.0 & 96.1 \\
\bottomrule
\end{tabular}

\label{tab:retrieval_results}
\end{table*}

\section{Experiment}

\subsection{Experimental Settings}
\paragraph{Implementation Details.} 
We employed ViT-H/14~\cite{vit} as the visual backbone and utilized RoBERTa-wwm-large~\cite{roberta} as the textual backbone to construct the Chinese-CLIP~\cite{yang2022chinese} and CLIP~\cite{radford2021learning} models. The experimental trials were executed utilizing a single NVIDIA RTX A6000 GPU, endowed with 48 gigabytes of memory. The fine-tuning process was implemented with a batch size of 32, a learning rate set to 5e-5, and trained for 3 epochs. We utilize Janus-Pro 7B~\cite{chen2025janus} for attribute extraction and caption generation, and Qwen-2.5VL~\cite{Qwen2.5-VL} for consistency evaluation. For the inference time, $C^3$ takes 2.9~s per caption, while Janus-Pro with the default prompt template takes 2.2~s per caption. 

\paragraph{Datasets and Evaluation Metrics.}We perform experiments on two publicly cross-modal retrieval datasets, MSCOCO~\cite{mscoco} and Flickr30K~\cite{30k}, as well as on a domain-specific cultural dataset CulTi~\cite{culti}, and a historical artifact dataset TimeTravel~\cite{timetravel}. MSCOCO comprises 123,287 images, each paired with five English captions, of which 113,287 are used for training, 5,000 for validation, and 5,000 for testing. The Flickr30K dataset contains 31,783 images with five English descriptions each, and we allocate 29,783 images for training and 1,000 each for validation and testing. Since our method augments captions based on the original captions, we use a one-to-one image-text pairing for retrieval. CulTi includes 5,726 Chinese image text pairs spanning two categories, silk artifacts and Dunhuang murals, with 4,008 pairs for training, 573 for validation, and 1,145 for testing. TimeTravel contains 10,250 expert-verified image–text pairs from 266 cultural groups across 10 regions, designed for historical artifact analysis and cultural context understanding. We split the dataset into 7,175 samples for training, 1,025 for validation, and 2,050 for testing. Retrieval performance is evaluated using Recall@$K$ ($R$@$K$), with $K \in \left\{1, 5, 10\right\}$.
\begin{figure*}
    \centering
    \includegraphics[width=1\linewidth]{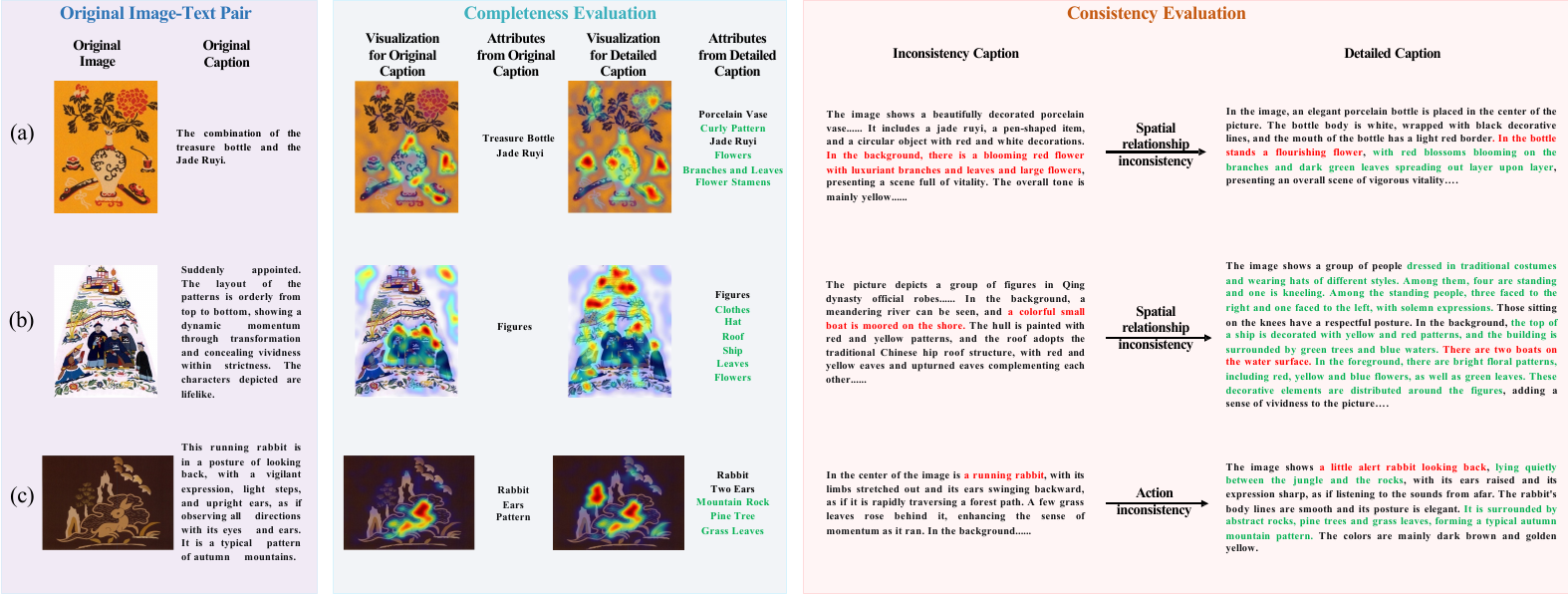}
    \caption{Case studies on completeness and consistency evaluation. The examples of captions and annotations are translated from Chinese to English for better understanding. \textcolor{red}{Red} denotes inaccurate or hallucinated descriptions;
    \textcolor{cctgreen}{Green} denotes missing details in the original caption.}
    \label{casestdudy}
\end{figure*}
\subsection{Main Results}
\paragraph{Zero-shot Retrieval.} As shown in Tab. \ref{tab:retrieval_results}. $C^3$ achieves competitive retrieval performance in zero-shot setting for  MSCOCO~\cite{mscoco} and Flickr30K~\cite{30k}, with accuracy exceeding 99\%. As text description is less pronounced in these datasets, augmentation brings limited improvement. However, $C^3$ significantly improves the original CLIP baseline ~\cite{radford2021learning} in both public datasets. On the more challenging CulTi dataset, $C^3$ outperforms LACLIP ~\cite{culti} by 18.2\% at $R$@$1$ in the zero-shot setting as shown in Tab. \ref{tab:culture_timetravel_results}. On the TimeTravel dataset, $C^3$ also surpasses direct prompting by an average of 2.8\% at $R$@$1$ across both retrieval directions, confirming its effectiveness on historical artifact retrieval. By validating both visual attributes and their semantic grounding, $C^3$ enhances description quality and enables superior cross-modal alignment and generalization, even without domain-specific fine-tuning. The shortfall on Flickr30K is mainly due to the CLIP-based backbone. Its pre-training is less effective for fine-grained phrase grounding required by Flickr30K.

\begin{table}[ht]
\centering
\renewcommand{\arraystretch}{1.2}
\setlength{\tabcolsep}{2.5pt}
\scriptsize
\caption{Retrieval performance on CulTi and TimeTravel test sets. For the direct prompt, we use the default prompt template provided by Janus-Pro.}
\begin{tabular}{c|ccc|ccc}
\toprule
\multirow{2}{*}{\centering \textbf{Methods}} & \multicolumn{3}{c|}{\textbf{Image-to-Text}} & \multicolumn{3}{c}{\textbf{Text-to-Image}} \\
& R@1 & R@5 & R@10 & R@1 & R@5 & R@10 \\
\midrule

\multicolumn{7}{c}{\textbf{CulTi (zero-shot)}} \\
LACLIP~\cite{culti} & 7.6 & 22.9 & 31.1 & 11.0 & 25.1 & 36.2 \\
Direct Prompt & 20.6 & 43.6 & 54.2 & 25.4 & 48.3 & 60.1 \\
\rowcolor{Magenta!5}\textbf{$C^3$ (Ours)} & \textbf{25.8} & \textbf{51.3} & \textbf{61.9} & \textbf{26.4} & \textbf{55.5} & \textbf{65.7} \\

\multicolumn{7}{c}{\textbf{CulTi (fine-tune)}} \\
LACLIP~\cite{culti} & 23.6 & 49.9 & 62.9 & 28.1 & 56.6 & 66.2 \\
Direct Prompt & 54.6 & 80.4 & 89.3 & 53.7 & 80.3 & 88.1 \\
\rowcolor{Magenta!5}\textbf{$C^3$ (Ours)} & \textbf{74.1} & \textbf{93.4} & \textbf{96.4} & \textbf{77.1} & \textbf{94.6} & \textbf{97.4} \\
\midrule

\multicolumn{7}{c}{\textbf{TimeTravel (zero-shot)}} \\
Original Caption & 10.6 & 26.0 & 35.3 & 9.8 & 24.9 & 33.8 \\
Direct Prompt & 21.0 & 40.8 & 50.5 & 17.5 & 35.3 & 45.0 \\
\rowcolor{Magenta!5}\textbf{$C^3$ (Ours)} & \textbf{24.0} & \textbf{44.6} & \textbf{54.8} & \textbf{20.0} & \textbf{38.3} & \textbf{47.8} \\

\multicolumn{7}{c}{\textbf{TimeTravel (fine-tune)}} \\
Original Caption & 21.3 & 49.8 & 65.6 & 21.0 & 48.3 & 64.6 \\
Direct Prompt & 42.7 & 70.4 & 81.4 & 43.7 & 71.0 & 80.7 \\
\rowcolor{Magenta!5}\textbf{$C^3$ (Ours)} & \textbf{46.6} & \textbf{75.2} & \textbf{86.9} & \textbf{48.4} & \textbf{76.9} & \textbf{86.3} \\
\bottomrule
\end{tabular}
\label{tab:culture_timetravel_results}
\end{table}

\begin{figure}[h!]
    \centering
    \includegraphics[width=1\linewidth]{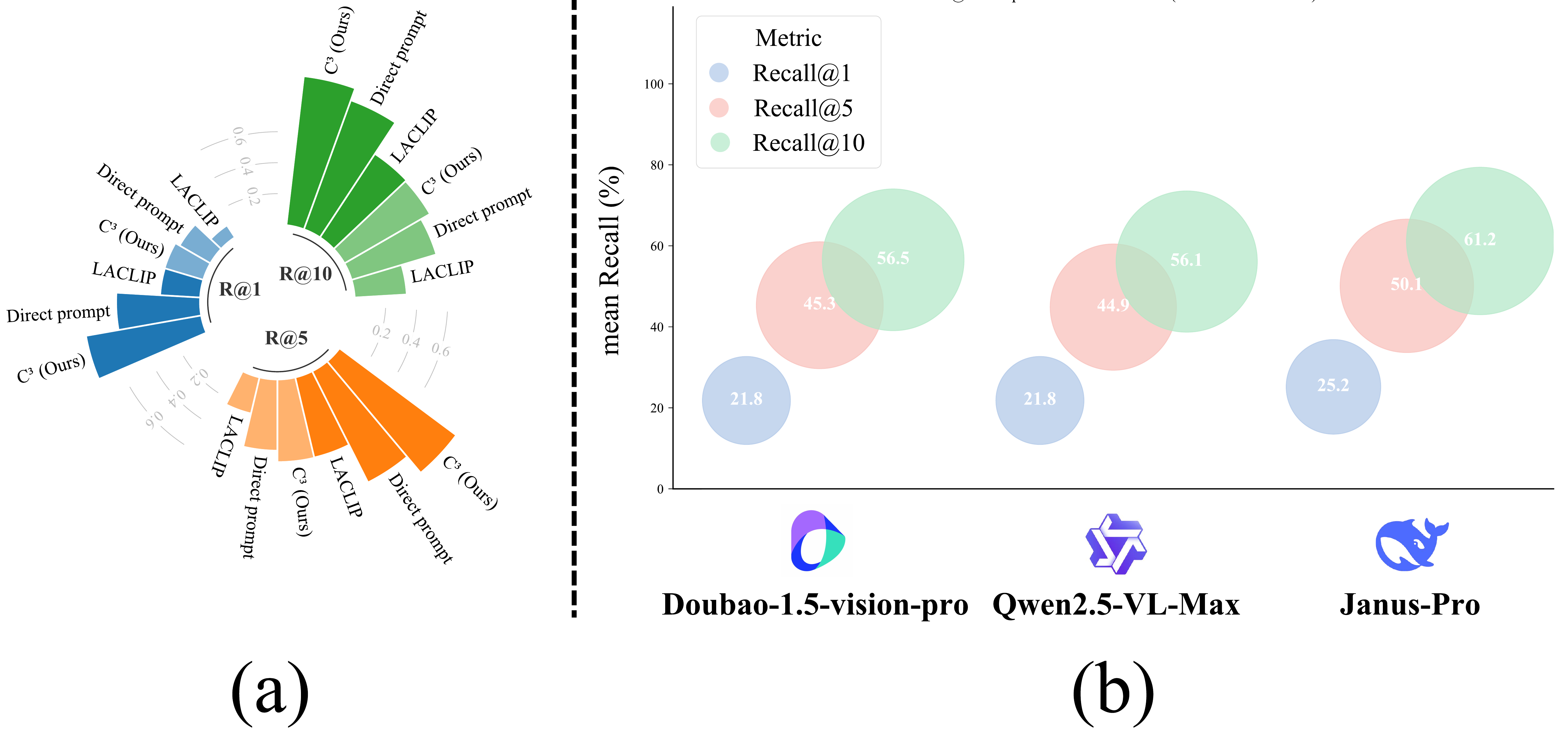}
    \caption{(a) Retrieval performance on the CulTi dataset under both zero-shot and fine-tune conditions; (b) Comparisons of zero-shot retrieval performance between different VLLMs in CulTi.}
    \label{fig:enter-label}
\end{figure}

\paragraph{Cultural Heritage Domain Retrieval.} 
As shown in Tab.~\ref{tab:culture_timetravel_results} and in Fig.~\ref{fig:enter-label} (a), the directly prompting-based method outperforms LACLIP~\cite{culti} in CulTi. Our approach further improves upon directly prompting by 19.5\% at $R$@$1$ for image-to-text retrieval in a fine-tuned setting. On the TimeTravel dataset, $C^3$ also surpasses direct prompting by 4.3\% at $R$@$1$ in the fine-tuned setting, highlighting its effectiveness on historical artifact retrieval. These results demonstrate the robustness of $C^3$ on challenging cultural data. By enabling attribute-level bidirectional evaluation, our method achieves fine-grained alignment between visual details and textual attributes, effectively addressing the incompleteness and ambiguity common in cultural captions. Besides, Fig.~\ref{fig:enter-label} (b) shows that $C^3$ achieves similar retrieval performance across different VLLM parameter sizes, indicating that $C^3$ is not dependent on model scale.


\subsection{Ablation studies}

\subsubsection{Importance of Completeness Evaluation}
We conduct an ablation study to demonstrate the effectiveness of the completeness evaluation as shown in Tab. \ref{tab:ablation_study} (b). To remove the completeness evaluation, we no longer extract attributes from augmentation, and each verification question is assigned an equal cost. Under these conditions, the mean Recall (mR) and sum of Recalls (Rsum) decrease by 7.5\% and 44.7\% on CulTi. 
This ablation highlights that completeness evaluation ensures that all visual regions and attributes are accounted for, thus significantly improving the representation of image content in the captions. Moreover, this step is particularly important in the cultural heritage domain, where visual attributes are diverse and frequently subtle, requiring precise completeness evaluation to support accurate retrieval.

\begin{table}[ht]
\centering
\caption{Ablation studies on three datasets.}
\label{tab:ablation_study}
\resizebox{\columnwidth}{!}{
\begin{tabular}{c|l|cc|cc|cc}
\toprule
\multirow{2}{*}{No.} & \multirow{2}{*}{Model Variant} 
                    & \multicolumn{2}{c|}{Flickr30k} 
                    & \multicolumn{2}{c|}{MSCOCO} 
                    & \multicolumn{2}{c}{CulTi} \\  
    &               & mR   & Rsum   & mR    & Rsum    & mR    & Rsum   \\
\midrule
(a) & \textbf{$C^3$ (Ours)}               & 95.5 & 572.8 & 93.9 & 563.3 & 88.8 & 532.8 \\
(b) & w/o Completeness Evaluation         & 94.5 & 567.0 & 90.5 & 542.8 & 81.3 & 488.1 \\ 
(c) & w/o CoT                             & 94.6 & 567.6 & 88.9 & 533.2 & 74.4 & 446.5 \\
(d) & w/o Consistency Evaluation          & 95.3 & 571.6 & 93.6 & 561.8 & 86.8 & 520.5 \\
\bottomrule
\end{tabular}
}
\end{table}


\subsubsection{Importance of Consistency Evaluation}
We further evaluate the necessity of consistency evaluation. In this case, we remove the Chain-of-Thought (CoT) and Markov decision-based VQA evaluation. First, we directly generate captions using simple prompting as shown in Tab. \ref{tab:ablation_study} (c). In this scenario, the mR and Rsum drop by 14.4\% and 86.3\%, respectively. We analyze that the model cannot decompose and reason complex attribute information, resulting in a significant drop in retrieval performance. 
Second, we apply CoT without consistency evaluation in Tab. ~\ref{tab:ablation_study} (d). In this case, mR and Rsum drop by 2.0\% and 12.3\%. Additionally, our consistency evaluation yields a dialog reasoning accuracy of 93.2\%, evidencing the reliability of the evaluator in guiding caption refinement. 
The lack of consistency evaluation increases the risk of hallucinations, especially when dealing with previously unseen cultural heritage concepts. Therefore, consistency evaluation is crucial in LLM-based data augmentation to prevent inaccuracies and ensure caption reliability in diverse cultural contexts.

\subsection{Case Study}
In this section, we illustrate the advantage of $C^3$ through three case studies in Fig. \ref{casestdudy}. These case studies focus on two typical problems that occurred during the caption augmentation processing, but could be addressed by our $C^3$ framework. Specifically, in Fig. \ref{casestdudy} (a), the original and LLM-generated captions mention only the bottle and ruyi but omit the flowers at the top. Our method adds descriptions for these flowers so retrieval can account for them. 
By covering every visual attribute, our augmented captions form more complete queries. Therefore, the retrieval model performs stronger alignment across all described elements. 
In Fig. \ref{casestdudy} (c), the rabbit’s action is unclear from the image alone, but the original caption uses the word “looking back”. This shows that the caption “running rabbit” is inaccurate. $C^3$ is able to identify these issues because the proposed consistency evaluation leverages completeness evaluation to find objective evidence from both the image and the text. Furthermore, our approach asks targeted questions at each stage of the CoT process, narrowing the scope of evaluation and enabling precise checking.

\section{Conclusion}
In this paper, we addressed the challenge of incomplete and unreliable textual annotations in cultural heritage retrieval through $C^3$, an LLM-driven data augmentation framework. By introducing a novel bidirectional validation approach and a coverage-based measure mechanism, our model ensures comprehensive textual descriptions. Furthermore, the integration of Chain-of-Thought prompting supervised by a Markov decision process effectively mitigates hallucination, significantly enhancing reliability. Empirical results validate our method's efficacy, achieving superior performance on cultural heritage retrieval tasks and demonstrating robust generalizability across benchmark datasets. Future work will explore extending this approach to additional modalities and further refining semantic alignment strategies for broader applicability.

\section{Limitations}
Although our proposed method $C^3$ has demonstrated promising performance on cultural heritage datasets, it still exhibits minor inaccuracies in generated captions. Upon manual inspection, we found occasional errors primarily arising from incomplete visual information, such as damaged murals or poor image quality, which can mislead the model’s assessment of content completeness. Additionally, the consistency evaluation requires multi-turn verification for difficult cultural images, and LLMs may forget context over multiple rounds, sometimes causing incorrect captions to be mistakenly accepted as correct. Future research will address these limitations by incorporating robustness to incomplete visual contexts and by investigating dynamic context optimization strategies for the evaluation process.


\section*{Acknowledgments}
This project is supported by the National Natural Science Foundation of China (Nos. 62436009, 62276258), the XJTLU Research Development Fund and Teaching Development Fund (Grant No. RDF-24-01-016, TDF23/24-R27-222), and the Suzhou Science and Technology Development Planning Programme (Grant No. ZXL2023176).                                                                                                                                                                                                                                                                                                                                                                                                                                                                                                                                                                                                                                                                                                                                                                                                                                                                                                                                                                                                                                                                                                                                                                                                                                                                                                                                                                                                                                                                                                                                                                                                                                                                                                                                                                                                                                                                                                                                                                                                                                                                                                                                                                                                                                                                                                                                                                                                                                                                                                                                                                                                                                                                                                                                                                                                                                                                                                                                                                                                                                                                                                                                                                                                                                                                                                                                                                                                                                                                                                                                                                                                                                                                                                                                                                                                                                                                                                                                                                                                                                                                                                                                                                                                                                                                                                                                                                                                                                                                                                                                                                                                                                                                                                                                                                                                                                                                                                                                                                                                                                                                                                                                                                                                                                                                                                                                                                                                                                                                                                                                                                                                                                                                                                                                                                                                                                                                                                                                                                                                                                                                                                                                                                                                                                                                                                                                                                                                                                                                                                                                                                                                                                                                                                                                                                                                                                                                                                                                                                                                                                                                                                                                                                                                                                                                                                                                                                                                                                                                                                                                                                                                                                                                                                                                                                                                                                                                                                                                                                                                                                                                                                                                                                                                                                                                                                                                                                                                                                                                                                                                                                                                                                                                                                                                                                                                                                                                           


\appendix

\bibliography{main}
\end{document}